\documentclass[10pt,twocolumn,letterpaper]{article}

\usepackage{iccv}
\usepackage{times}
\usepackage{epsfig}
\usepackage{graphicx}
\usepackage{amsmath}
\usepackage{amssymb}
\usepackage{multirow}
\usepackage{mathtools}

\usepackage[pagebackref=true,breaklinks=true,letterpaper=true,colorlinks,bookmarks=false]{hyperref}

\iccvfinalcopy % *** Uncomment this line for the final submission

\def\iccvPaperID{1483} % *** Enter the CVPR Paper ID here

\ificcvfinal\fi
\begin{document}

%%%%%%%%% TITLE

\title{ Learning Robust Visual-Semantic Embeddings }

\author{Yao-Hung Hubert Tsai\,\,\,\,\,\,\,\,\, Liang-Kang Huang \,\,\,\,\,\,\,\,\, Ruslan Salakhutdinov \\
School of Computer Science, Machine Learning Department, Carnegie Mellon University\\
{\tt yaohungt@cs.cmu.edu}\,\,\,\, {\tt liangkah@andrew.cmu.edu}\,\,\,\, {\tt rsalakhu@cs.cmu.edu}
}

\maketitle
%\thispagestyle{empty}

%%%%%%%%% ABSTRACT
\begin{abstract}

Many of the existing methods for learning joint embedding of images and text use only supervised information from paired images and its textual attributes. Taking advantage of the recent success of unsupervised learning in deep neural networks, we propose an end-to-end learning framework that is able to extract more robust multi-modal representations across domains. The proposed method combines representation learning models (i.e., auto-encoders) together with cross-domain learning criteria (i.e., Maximum Mean Discrepancy loss) to learn joint embeddings for semantic and visual features. A novel technique of unsupervised-data adaptation inference is introduced to construct more comprehensive embeddings for both labeled and unlabeled data. We evaluate our method on Animals with Attributes and Caltech-UCSD Birds 200-2011 dataset with a wide range of applications, including zero and few-shot image recognition and retrieval, from inductive to transductive settings. 
Empirically, we show that our framework improves over the current state of the art on many of the considered tasks.

\end{abstract}

%%%%%%%%% BODY TEXT
%!TEX root = ./egpaper_for_review.tex
\vspace{-2mm}
\section{Introduction}
\vspace{-1mm}
Over the past few years, due to the availability of large amount of data and the advancement of the training techniques, learning effective and robust representations directly from images or text becomes feasible~\cite{alex2012imagenet,mikolov2013distributed,pennington2014glove}. These learned representations have facilitated a number of high-level tasks, such as image recognition \cite{simonyan2014very}, sentence generation \cite{kiros2015skip}, and object detection \cite{ren2015faster}. Despite useful representations being developed for specific domains, learning more comprehensive representations across different data modalities remains challenging. In practice, more complex tasks, such as image captioning \cite{oriol2015show} and image tagging \cite{li2015zero} often involve data from different modalities (i.e., images and text). 
Additionally, the learning process would be faster, requiring fewer labeled examples, and hence more scalable to handling a large number of categories if we could transfer cross-domain knowledge more effectively \cite{frome2013devise}. 
This motivates learning multi-modal embeddings. In this paper, we consider learning robust joint embeddings across visual and textual modalities in an end-to-end fashion under zero and few-shot setting. 

\begin{figure}[t!]
\includegraphics[width=0.46\textwidth]{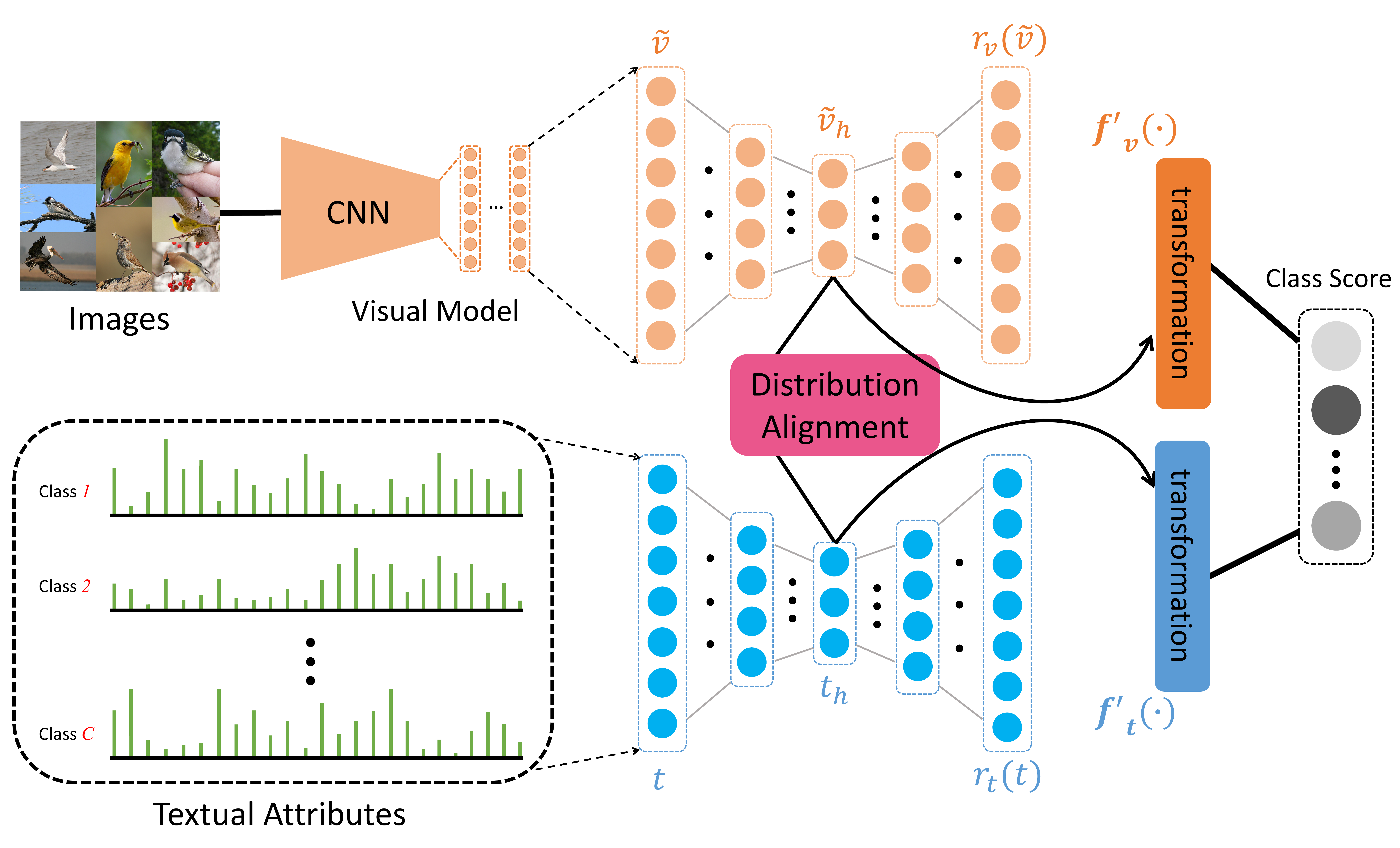}
\caption{\footnotesize Illustration of our proposed ReViSE (Robust sEmi-supervised Visual-Semantic Embeddings) architecture.}
\label{fig:prop_method}
\vspace{-0.1in}
\end{figure}

Zero-shot learning aims at performing specific tasks, such as recognition and retrieval of novel classes, when no label information is available during training \cite{jayaraman2014zero}. On the other hand, few-shot learning enables us to have few labeled examples in our of-interest categories \cite{salakhutdinov2012one}. In order to compensate the missing information under the zero and few-shot setting, the model should learn to associate novel concepts in image examples with textual attributes and transfer knowledge from training to test classes. A common strategy for deriving the visual-semantic embeddings is to make use of images and textual attributes in a supervised way \cite{socher2013zero,akata2015evaluation,xian2016latent,zhang2015zero,zhang2016zero1,lei2015predicting,changpinyo2016synthesized}. Specifically, one can learn transformations of images and textual attributes under the objective that the transformed visual and semantic vectors of the same class should be similar in the joint embeddings space. Despite good performance, this common strategy basically boils down to a supervised learning setting, learning from labeled or paired data only. In this paper, we show that to learn better joint embeddings across different data modalities, it is beneficial to combine supervised and unsupervised learning from both labeled and unlabeled data.

Our contributions in this work are as follows. First, to extract meaningful feature representations from both labeled and unlabeled data, one possible option is to train an auto-encoder~\cite{radford2015dcgan,bengio2013generalized}. 
In this way, instead of learning representations directly to align the visual and textual inputs, we choose to learn representations %transformations on the hidden representations 
in an auto-encoder using reconstruction objective. Second, we impose a cross-modality distribution matching constraint to require the embeddings learned by the visual and textual auto-decoders to have similar distributions. By minimizing the distributional mismatch between visual and textual domain, we show improved performance on recognition and retrieval tasks. Finally, to achieve better adaptation on the unlabeled data, we perform a novel unsupervised-data adaptation inference technique. We show that by adopting this technique, the accuracy increases significantly not only for our method but also for many of the existing other models. Fig. \ref{fig:prop_method} illustrates our overall end-to-end differentiable model.

To summarize, our proposed method successfully combines supervised and unsupervised learning objectives, and learns from both labeled and unlabeled data to construct joint embeddings of visual and textual data. We demonstrate improved performance on Animals with Attributes ($\mathsf{AwA}$) \cite{lampert2014attribute} and Caltech-UCSD Birds 200-2011 \cite{WelinderEtal2010} datasets on both image recognition and image retrieval tasks under zero and few-shot setting.

\section{Related Work}
\label{sec:related}
In this section, we provide an overview of learning multi-modal embeddings across visual and textual domain. 
\vspace{.05in}

\hspace{-5mm} {\bf Zero and Few-Shot Learning}
\vspace{.05in}

Zero-shot \cite{changpinyo2016synthesized, akata2016multi, akata2015evaluation} and few-shot learning \cite{fei2006oneshot,salakhutdinov2013learning,lake2013one} are related problems, but somewhat different in the setting of the training data. While few-shot learning aims to learn specific classes through one or few examples, zero-shot learning aims to learn even when no examples of the classes are presented. In this setting, zero-shot learning should rely on the side information provided by other domains. In the case of image recognition, this often comes in the form of textual descriptions. Thus, the focus of zero-shot image recognition is to derive joint embeddings of visual and textual data, so that the missing information of specific classes could be transferred from the textual domain. 

Since the relation between raw pixels and text descriptions is non-trivial, most of the previous work relied on learning the embeddings through a large amount of data. Witnessing the success of deep learning in extracting useful representations, much of the existing work mostly applies deep neural networks to first transform raw pixels and text into more informative representations, followed by using
various techniques to further identify the relation between them. For example, Socher {\em et al.} \cite{socher2013zero} used deep architectures \cite{huang2012improving} to learn representations for both images and text, and then used a Bayesian framework to perform classification. Norouzi {\em et al.} \cite{norouzi2013zero} introduced a simple idea that treated classification scores output by the deep network \cite{alex2012imagenet} as weights in convex combination of word vectors. Fu {\em et al.} \cite{fu2015transductive} proposed a method that learns projections 
from low-level visual and textual features to form a hypergraph in the embedding space and performed label propagation for recognition. A number of similar methods learn transformations from input image representations to the semantic space for the recognition or retrieval purposes \cite{akata2015evaluation,zhang2015zero,akata2016multi,xian2016latent,changpinyo2016synthesized,zhang2016zero1,zhang2016zero2,palatucci2009zero,romera2015embarrassingly,bucher2016improving,guo2016transductive}.

%Instead of designing algorithms to work out, the joint embeddings from the output of deep models, other approaches aimed to learn the whole task with deep models in an end-to-end fashion. 
A number of recent approaches also attempt to learn the entire task with deep models in an end-to-end fashion.
Frome {\em et al.} \cite{frome2013devise} constructed a deep model that took visual embeddings extracted by CNN~\cite{alex2012imagenet} and word embeddings as input, and trained the model with the objective that the visual and word embeddings of the same class should be well aligned under linear transformations. Ba {\em et al.} \cite{lei2015predicting} predicted the output weights of both the convolutional and fully connected layers in a deep convolutional neural network. % in the supervision of text features. 
Instead of using textual attributes or word embeddings model, Reed {\em et al.} \cite{reed2016learning} proposed to train a neural language model directly from raw text with the goal of encoding only the relevant visual concepts for various categories.

\hspace{-5mm} {\bf Visual and Semantic Knowledge Transfer}

Liu {\em et al.} \cite{liu2015multi} developed multi-task deep visual-semantic embeddings model for selecting video thumbnails based on side semantic information (i.e., title, description, and query). By incorporating knowledge about objects similarities between visual and semantic domains, Tang {\em et al.} \cite{tang2016large} improved object detection in a semi-supervised fashion. Kottur {\em et al.} \cite{kottur2015visual} proposed to learn visually grounded word embeddings (vis-w2v) and showed improvements over text only word embeddings (word2vec) on various challenging tasks. Reed {\em et al.} designed a text-conditional convolutional GAN architecture to synthesize an image from text. Recently, Wang {\em et al.} \cite{wang2015learning} introduced structure-preserving constraints in learning joint embeddings of images and text for image-to-sentence and sentence-to-image retrieval tasks.

%\hspace{-5mm} {\bf Domain Adaptation}

%Domain adaptation aims at adapting models from source to target domain given the scenario that cross-domain feature distributions are distinct \cite{pan2011domain}. While the exact problem setting might be different, the spirit of the problem is basically for learning modal-invariant embeddings. For example, finding a joint visual-semantic embedding given images and text data can be viewed as Domain Adaptation in general. Therefore, this suggests that some of the domain adaptation techniques might be applicable to our target problem. Among several techniques, we adopt Maximum Mean Discrepancy (MMD) \cite{gretton2006kernel} criterion as a distributions matching constraints in our proposed method. Examples of using MMD as a loss for training deep models can be found in \cite{long2015learning,tzeng2015simultaneous,tzeng2014deep}.

\hspace{-5mm} {\bf Unsupervised Multi-modal Representations Learning}

One of our key contributions is to effectively combine supervised and unsupervised learning tasks for learning multi-modal embeddings. This is inspired and supported by several previous works that provided evidence of how unsupervised learning tasks could benefit cross-modal feature learning. 

Ngiam {\em et al.} \cite{ngiam2011multimodal} proposed various models based on Restricted Boltzmann Machine, Deep Belief Network, and Deep Auto-encoder to perform feature learning over multiple modalities. The derived multi-modal features demonstrated an improved performance over single-modal features on the audio-visual speech classification tasks. Srivastava and Salakhutdinov \cite{srivastava2012multimodal} developed a Multimodal Deep Boltzmann Machine for fusing together multiple diverse modalities even when some of them are absent. Providing inputs of images and text, their generative model manifested noticeable performance improvement on classification and retrieval tasks.

\section{Proposed Method}

First, we define the problem setting and the corresponding notation. 
Let $\mathbf{V_{tr}} = \{v_{i}^{(tr)}\}_{i=1}^{N_{tr}}$ denote labeled training images from $C_{tr}$ classes, $\mathbf{V_{ut}} = \{v_{i}^{(ut)}\}_{i=1}^{N_{ut}}$ denote unlabeled training images from $C_{ut}$ possibly different classes, and $\mathbf{V_{te}} = \{v_{i}^{(te)}\}_{i=1}^{N_{te}}$ denote test images from $C_{te}$ novel classes. 
For each class, following \cite{zhang2015zero,zhang2016zero1,zhang2016zero2,akata2015evaluation,xian2016latent,changpinyo2016synthesized}, 
its textual attributes are either provided from human annotated attributes \cite{lampert2014attribute} or learned from unsupervised text corpora (Wikipedia) \cite{pennington2014glove}. 
We denote these class-specific textual attributes as $\mathbf{T_{tr}} = \{t_{c}^{(tr)}\}_{c=1}^{C_{tr}}$, $\mathbf{T_{ut}} = \{t_{c}^{(ut)}\}_{c=1}^{C_{ut}}$, and $\mathbf{T_{te}} = \{t_{c}^{(te)}\}_{c=1}^{C_{te}}$ for labeled training, unlabeled training, and test classes, respectively. 

Under zero-shot setting, 
our goal is to predict labels of the test images coming from 
{\it novel}, previously unseen, classes 
given textual attributes.
%labels of the test images are unknown. Our goal is to predict these unknown labels given textual attributes. 
That is, for a given test image $v_{i}^{(te)}$, its label is determined by 
\vspace{-2mm}
\begin{equation}
\label{eq:pred_test_label}
\vspace{-2mm}
\underset{c\in\{1,...,C_{te}\}}{\mathrm{arg\,max}} P_{\theta}\left ( c\,\middle|t_c^{(te)}, v_i^{(te)} \right ),
\end{equation}
where $\theta$ denotes model parameters. We will also consider a few-shot learning, where a few labeled training images are available in each of the test classes. In the following, we omit the model parameters $\theta$ for brevity.

\subsection{Basic Formulation}

The goal of learning multi-modal embeddings can be formulated as learning transformation functions $f_v$ and $f_t$, such that given an image $v$ and a textual attribute $t$, $f_v(v)$ should be similar to $f_t(t)$ if $v$ and $t$ are of the same class. Much of the previous work for learning multi-modal embeddings can be generalized to this formulation. For instance, in Cross-Modal Transfer (CMT) \cite{socher2013zero}, $f_v(\cdot)$ can be viewed as a pre-defined feature extraction model followed by a two-layer neural network, while $f_t(\cdot)$ is set to an identity matrix. To be more specific, \cite{socher2013zero} aim at learning a non-linear projection directly from visual features to semantic word vectors. 

Over the past few years, deep architectures have been shown to learn useful representations that could embed high-level semantics for both visual and textual data. This gives rise to the attempts of applying successful deep architectures to learn $f_v(\cdot)$ and $f_t(\cdot)$. For example, DeViSE \cite{frome2013devise} designed $f_v(\cdot)$ as a CNN model followed by a linear transformation matrix. On the other hand, they adopted the well known skip-gram text modeling architecture \cite{mikolov2013distributed} for learning $f_t(\cdot)$ from raw text on Wikipedia. It is worth noting that, to further take advantage of previous success, these deep models are often pre-trained on large datasets where they have shown to learn effective representations.

\begin{figure}[t!]
\centering
\includegraphics[width=0.3\textwidth]{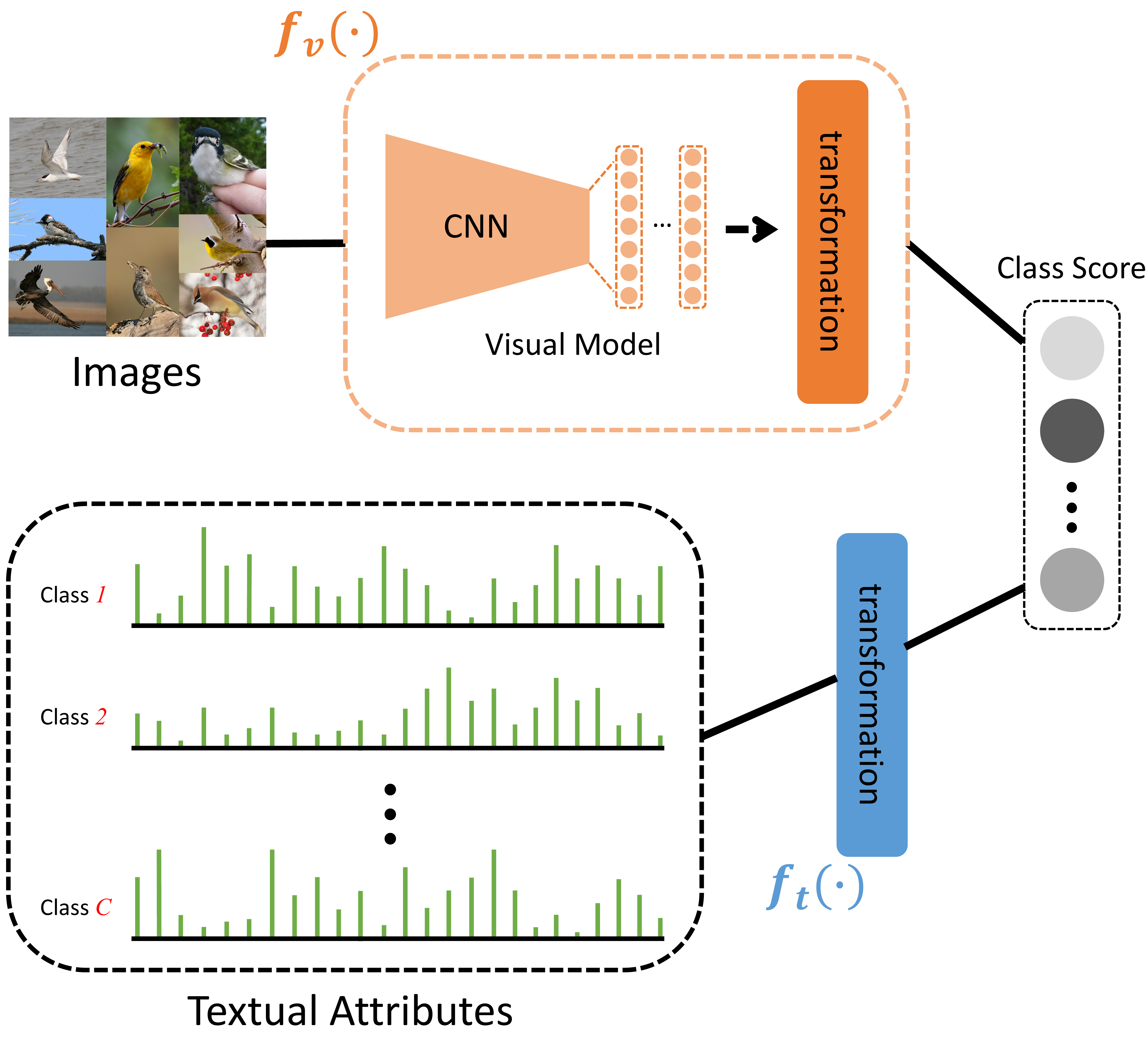}
\caption{\footnotesize Illustration of traditional visual-semantic embeddings model.}
\label{fig:basic}
\vspace{-3mm}
\end{figure}

Figure \ref{fig:basic} shows the basic formulation of the visual-semantic embeddings model. 
Our method is built on top of this basic architecture by adding additional components as well as modifying existing ones.

\subsection{Reconstructing Features from Auto-Encoder}

Although the basic architecture provides a way to utilize label information during training, the learning process could further benefit if unlabeled data are provided at the same time. To be more specific, we propose to combine supervised and unsupervised learning objectives together by incorporating auto-encoders \cite{bengio2009learning} for both image and text data. Typical setting of auto-encoders consists of a symmetric encoder-decoder architecture, with the hidden representations in the middle being compact representations that could be used to reconstruct the original input data. In our model, the auto-encoders are added after the image and text data are processed by the pre-trained networks. For learning visual embeddings, we use contractive auto-encoder proposed by \cite{rifai2011contractive}, which is able to learn more robust visual codes for the images of same class. Given a visual feature vector $\tilde{v}$, the contractive auto-encoder maps $\tilde{v}$ to a hidden representation $\tilde{v}_h$, and seeks to reconstruct $\tilde{v}$ from $\tilde{v}_h$. Let us denote the reconstructed vector by $r_v(\cdot)$. Model parameters are thus learned by minimizing the regularized reconstruction error
\begin{equation}
\label{eq:v_auto} 
    \mathcal{L}_v = \frac{1}{N_{tr}} \sum_{i=1}^{N_{tr}}  \lVert \tilde{v}_i-r_v(\tilde{v}_i) \rVert^2 + \gamma \lVert J(\tilde{v}_i) \rVert_F^2, 
\end{equation}
%   \mathcal{L}_v = \frac{1}{N_{tr}+N_{te}} \sum_{i=1}^{N_{tr}+N_{te}}  \lVert \tilde{v}_i-r_v(\tilde{v}_i) \rVert^2 + \gamma \lVert J(\tilde{v}_i) \rVert_F^2, 
where $J(\cdot)$ is the Jacobian matrix~\cite{rifai2011contractive}.

On the other hand, for a given semantic feature vector or textual attribute $t$, we use a vanilla auto-encoder to first encode and then reconstruct from its hidden representation~$t_h$. 
We hence minimize the reconstruction error
\vspace{-1mm}
\begin{equation} 
\label{eq:t_auto}
    \mathcal{L}_t  = \frac{1}{C_{tr}} \sum_{c=1}^{C_{tr}}  \lVert t_c-r_t(t_c) \rVert^2.
\end{equation}
%   \mathcal{L}_t  = \frac{1}{C_{tr}+C_{te}} \sum_{i=1}^{C_{tr}+C_{te}}  \lVert t_i-r_t(t_i) \rVert^2.
Combining \eqref{eq:v_auto} and \eqref{eq:t_auto} gives us the reconstruction loss
\vspace{-1mm}
\begin{equation} 
\label{eq:recon}
    \mathcal{L}_{reconstruct} = \mathcal{L}_v + \mathcal{L}_t.
\end{equation}

In practice, if we have access to a large unlabeled set or a set of test inputs (without labels), we
can easily incorporate them into the reconstruction loss. 
In our experimental results, we find that with the visual and textual auto-encoders, 
image and text data are transformed into visual and textual embeddings with more meaningful 
representations. In order to further transfer the knowledge across modalities, we impose discriminative constraints on the hidden representations ($\tilde{v}_h$ and $t_h$) learned by these auto-encoders, as we discuss next.

\subsection{Cross-Modality Distributions Matching}
{
\label{subsec:mmd}

Distributions matching technique has been proven to be effective for transferring knowledge from one modality to another \cite{pan2010survey,long2015learning,hubert2016learning}. A common nonparametric way to analyze and compare distributions is to use Maximum Mean Discrepancy (MMD) \cite{gretton2006kernel} criterion. We can view MMD as a two-sample test on $\tilde{v}_h$ and $t_h$, and thus its loss can be formulated as
\begin{equation}
\label{eq:mmd} 
    \mathcal{L}_{MMD} = \lVert \mathbf{E}_{p}[\phi(\tilde{v}_h)] - \mathbf{E}_{q}[\phi(t_h)] \rVert_{\mathcal{H}_k}^2, 
\end{equation}
where $p$, $q$ are the distributions of visual and textual embeddings (i.e., $\tilde{v}_h \sim p$ and $t_h \sim q$), $\phi$ is the feature map with canonical form $\phi(\mathbf{x}) = k(\mathbf{x}, \cdot)$, and $\mathcal{H}_k$ is the reproducing kernel Hilbert space (RKHS) endowed with a characteristic kernel $k$. Note that the kernel in the MMD criterion must be a universal kernel, and thus we empirically choose a Gaussian kernel: % in our architecture:
% Gaussian kernel can be expressed as
\begin{equation}
\label{eq:rbf}
    k(\mathbf{x}, \mathbf{x}') = \mathrm{exp}\left( - \kappa \left\| \mathbf{x} - \mathbf{x}'\right\|^2 \right).
\end{equation}

We can now minimize the MMD criterion between visual and textual embeddings by minimize eq.~\eqref{eq:mmd}. This can be further regarded as shrinking the gap between information across two data modalities. In our experiments, we find that the MMD loss helps improve model performance on both recognition and retrieval tasks in zero and few-shot setting.

}

\subsection{Learning}
\label{subsec:learning}
{
After we derive the hidden representations $\tilde{v}_h$ and $t_h$, the transformation functions $f_v(\cdot)$ and $f_t(\cdot)$ can be reformulated as 
\begin{equation}
\label{eq:reform}
f_v(v) = f'_v(\tilde{v}_h) \,\, \mathrm{and} \,\, f_t(t) = f'_t(t_h),
\end{equation}
where $f'_v(\cdot)$ and $f'_t(\cdot)$ are the mapping functions from the hidden representations to the visual and textual output.

To leverage the supervised information from labeled training images $\mathbf{V_{tr}}$ and the corresponding textual attributes $\mathbf{T_{tr}}$, we minimize the binary prediction loss:
\begin{equation}
\label{eq:supervised}
    \mathcal{L}_{supervised} = -\frac{1}{N_{tr}} \sum_{i=1}^{N_{tr}} \sum_{c=1}^{C_{tr}} I_{i,c} \left \langle f'_v(\tilde{v}_{h,i}^{(tr)}), f'_t(t_{h,c}^{(tr)}) \right \rangle , 
\end{equation}
where $I_{i,c}$ indicates a $\{0,1\}$ encoding of positive and negative classes and $ \left \langle \cdot \right \rangle$ denotes a dot-product. It is worth noting that we can adopt different loss functions, including binary cross-entropy loss or multi-class hinge loss.
However, empirically, using the simple binary prediction loss results in the best performance 
of our model.

Similar to eq. \eqref{eq:supervised}, we adopt the binary prediction loss for unlabeled training images $\mathbf{V_{ut}}$ and the attributes $\mathbf{T_{ut}}$:
\begin{equation}
\label{eq:test_adaptation}
    \mathcal{L}^{unsup}_{unlab} = -\frac{1}{N_{ut}} \sum_{i=1}^{N_{ut}} \sum_{c=1}^{C_{ut}} \widehat{I}_{i,c}^{(ut)} \left \langle f'_v(\tilde{v}_{h,i}^{(ut)}), f'_t(t_{h,c}^{(ut)}) \right \rangle , 
\end{equation}
where 
\begin{equation}
\label{eq:testIic}
\widehat{I}_{i,c}^{(ut)} = \left\{\begin{matrix}
1 & \mathrm{if}\,\,c = \underset{c\in\{1,...,C_{ut}\}}{\mathrm{arg\,max}} \left \langle f'_v(\tilde{v}_{h,i}^{(ut)}), f'_t(t_{h,c}^{(ut)}) \right \rangle\\ 
 0 & \mathrm{otherwise.}
\end{matrix}\right.
\end{equation}

We refer to eq. \eqref{eq:test_adaptation} as {\em unsupervised-data adaptation inference}, which acts as a self-reinforcing strategy using the unsupervised data with unknown labels. The intuition is that by minimizing eq. \eqref{eq:test_adaptation}, we can further adapt our unlabeled data into the learning of $f'_v(\cdot)$ and $f'_t(\cdot)$ based on the empirical predictions. The choice of $\lambda$ does influence its effectiveness. However, we find that setting $\lambda = 1.0$ works quite well for many methods we considered in this work.

%In our experimental results we find that this {\em unsupervised-data adaptation inference} technique significantly improves model performance.

%In another viewpoint, given a test image $v_i^{(te)}$, eq.~\eqref{eq:pred_test_label} can be rewritten through MAP inference:
%\begin{equation}
%\label{eq:MAP}
%\underset{c\in\{1,...,C_{te}\}}{\mathrm{arg\,max}} P_{\theta}\left ( c \right ) P_{\theta} \left ( v_i^{(te)}\, ,  t_c^{(te)} \middle| c \right ),
%\end{equation}
%where $P_{\theta}\left ( c \right )$ denotes the prior uniform distribution and $P_{\theta} \left ( v_i^{(te)}\, ,  t_c^{(te)} \middle| c \right )$ is the 
%conditional probability parametrized by our model $\theta$. To be more specific, if we take 
%\begin{equation}
%\label{eq:condition}
%\mathrm{log}\,P_{\theta} \left ( v_i^{(te)}\, ,  t_c^{(te)} \middle| c \right ) \propto \widehat{I}_{i,c}^{(te)} \left \langle f'_v(\tilde{v}_{h,i}^{(te)}), f'_t(t_{h,c}^{(te)}) \right \rangle, 
%\end{equation}
%then optimizing eq. \eqref{eq:MAP} is the same as optimizing eq. \eqref{eq:test_adaptation}. This can be viewed as the generative model that generates images $\tilde{v}_{h,i}^{(te)}$ and attributes $t_c^{(te)}$ given the assigned label $c$ under our model.

In sum, our model learns by minimizing the total loss from both 
supervised and unsupervised objectives:
\begin{equation}
\label{eq:all}
        \mathcal{L}_{Total} = \mathcal{L}_{supervised} + \alpha  \mathcal{L}_{unsupervised},
\end{equation}
where
\begin{equation}
\label{eq:unsup}
\begin{split}
        \mathcal{L}_{unsupervised} = \mathcal{L}_{reconstruct} + \lambda \mathcal{L}^{unsup}_{unlab} + \beta  \mathcal{L}_{MMD}, 
\end{split}
\end{equation}
with $\alpha$, $\lambda$, and $\beta$ representing the trade-off parameters for different components. Note that we can also view the unsupervised objective here as a regularizer for learning more robust visual and textual representations (see Figure~\ref{fig:prop_method} for our overall model architecture).

Before computing the loss, we find it useful to perform~$\ell_2$ normalization on the output scores $f_v(v)$ and $f_t(t)$ along the batch-wise direction. It can be viewed as a mixture of Batch Normalization $\cite{ioffe2015batch}$ and Layer Normalization $\cite{ba2016layer}$. The idea is simple, we encourage the competence between different instances in the data batch, rather than across different categories.

}

%------------------------------------------------------------------------
%!TEX root = ./egpaper_for_review.tex
\section{Experiments}
\label{sec:exp}
{
In the experiments, we denote our proposed method as ReViSE (Robust sEmi-supervised Visual-Semantic Embeddings). Extensive experiments on zero and few-shot image recognition and retrieval tasks are conducted using two benchmark datasets: Animals with Attributes ($\mathsf{AwA}$) \cite{lampert2014attribute} and Caltech-UCSD Birds 200-2011 ($\mathsf{CUB}$) \cite{WelinderEtal2010}. $\mathsf{CUB}$ is a fine-grained dataset in which the objects are both visually and semantically very similar, while $\mathsf{AwA}$ is a more general concept dataset. We use the same $\textit{training (+validation)/ test}$ splits as in \cite{akata2015evaluation,xian2016latent}. Table \ref{tbl:dataset_stat} lists the statistics of the datasets.

\begin{table}[t!]
\centering
\caption{\footnotesize The statistics of $\mathsf{AwA}$ and $\mathsf{CUB}$ datasets. Images and classes are disjoint across $\textit{training(+ validation)}$ / $\textit{test}$ split.}
\vspace{1mm}
\scalebox{0.8}
{
\begin{tabular}{|c||cc||cc|}
\hline
\multirow{2}{*}{$\mathsf{Dataset}$} & \multicolumn{2}{|c||}{$\textit{training (+ validation)}$} & \multicolumn{2}{c|}{$\textit{test}$}    \\ 
                         & \# of images          & \# of classes         & \# of images & \# of classes \\ \hline \hline
$\mathsf{AwA}$ \cite{lampert2014attribute}                     & 24293                 & 40                    & 6180         & 10            \\$\mathsf{CUB}$ \cite{WelinderEtal2010}                     & 8855                  & 150                   & 2931         & 50            \\ \hline
\end{tabular}
}
\label{tbl:dataset_stat}
\vspace{-4mm}
\end{table}

To verify the performance of our method, we consider two state-of-the-art deep-embeddings methods: CMT \cite{socher2013zero} and DeViSE \cite{frome2013devise}. CMT and DeViSE can be viewed as a special case of our proposed method with $\alpha = 0$ (without using unsupervised objective in eq.~\eqref{eq:all}). The difference between them is that DeViSE learns a nonlinear transformation on raw visual images and textual attributes for the alignment purpose, while CMT only learns the nonlinear transformation from visual to semantic embeddings. %Generally speaking, all of these methods aim to learn modality-invariant embeddings across visual and textual space for recognition or retrieval.

We choose GoogLeNet \cite{szegedy2015going} as the CNN model in DeViSE, CMT, and our architecture. For the textual attributes of classes, we consider three alternatives: human annotated attributes ($\mathit{att}$) \cite{lampert2014attribute}, Word2Vec attributes ($\mathit{w2v}$) \cite{mikolov2013distributed}, and Glove attributes ($\mathit{glo}$) \cite{pennington2014glove}. $\mathit{att}$ are continuous attributes judged by humans: $\mathsf{CUB}$ contains 312 attributes and $\mathsf{AwA}$ contains 85 attributes. $\mathit{w2v}$ and $\mathit{glo}$ are unsupervised methods for obtaining distributed text representations of words. We use the pre-extracted Word2Vec and Glove vectors from Wikipedia provided by \cite{akata2015evaluation,xian2016latent}. Both $\mathit{w2v}$ and $\mathit{glo}$ are 400-dim. features.

\subsection{Network Design and Training Procedure}
\label{subsec:train_pro}
{
	Please see Supplementary for the design details of ReViSE and its parameters. Note that we report results averaged over $10$ random trials.

}

\begin{table}[t!]
\centering
\caption{\footnotesize Zero-shot recognition using top-1 classification accuracy (\%). $\textit{att}$ attributes are used to describe each category.}
\vspace{1mm}
\scalebox{0.72}
{
\begin{tabular}{|c||ll||ll||c|}
\hline
$\mathsf{Dataset}$ & \multicolumn{2}{c||}{$\mathsf{AwA}$}                                                      & \multicolumn{2}{c||}{$\mathsf{CUB}$}                                                     & average                          \\
                    $\textit{recognition for}$    & \multicolumn{1}{c}{$\mathbf{V_{ut}}$} & \multicolumn{1}{c||}{$\mathbf{V_{te}}$} & \multicolumn{1}{c}{$\mathbf{V_{ut}}$} & \multicolumn{1}{c||}{$\mathbf{V_{te}}$} & top-1 acc. \\ \hline\hline
\multicolumn{6}{|c|}{using only labeled training data}                                                                                                                                                                                     \\ \hline \hline
DeViSE \cite{frome2013devise}                         &        60.8        &          63.0         &       38.9       &              36.8            &        49.9                       \\
CMT \cite{socher2013zero}                 &    59.3      &    61.6   &         41.1   &        40.6         &    50.7                 \\
ReViSE$^a$        &     60.3      &      61.2        &    46.4      &        45.0      &        53.2              \\
ReViSE$^b$        &     64.5           &     65.0         &     49.6         &              47.3            &      56.6               \\ \hline \hline
\multicolumn{6}{|c|}{using both labeled and unlabeled training data}   \\ \hline \hline
DeViSE* \cite{frome2013devise}             &   76.0        &       63.7       &     37.2    &          36.2          &    53.3             \\
CMT* \cite{socher2013zero}       &    77.8     &       58.5     &     39.9      &              39.8          &   54.0              \\
ReViSE$^c$        &   64.7         &     67.6        &   52.2     &    48.2   &       58.2        \\
ReViSE       &    {\bf 78.0  }      &   {\bf 68.6   }   &     {\bf 56.6   }   &  {\bf 49.6    }  &        {\bf 63.2}           \\ \hline
\end{tabular}
}
\label{tbl:induc_zero_recog}
\vspace{-1mm}
\end{table}

\begin{table}[t!]
\centering
\caption{\footnotesize Zero-shot retrieval using mean Average Precision (mAP) (\%). $\textit{att}$ attributes are used to describe each category.}
\vspace{1mm}
\scalebox{0.72}
{
\begin{tabular}{|c||ll||ll||c|}
\hline
$\mathsf{Dataset}$ & \multicolumn{2}{c||}{$\mathsf{AwA}$}                                                      & \multicolumn{2}{c||}{$\mathsf{CUB}$}                                                     & average                          \\
                    $\textit{retrieval for}$    & \multicolumn{1}{c}{$\mathbf{V_{ut}}$} & \multicolumn{1}{c||}{$\mathbf{V_{te}}$} & \multicolumn{1}{c}{$\mathbf{V_{ut}}$} & \multicolumn{1}{c||}{$\mathbf{V_{te}}$} & mAP \\ \hline\hline
\multicolumn{6}{|c|}{using only labeled training data}                                                                                                                                                                                     \\ \hline \hline
DeViSE \cite{frome2013devise}                         &        60.5        &      61.6       &         32.6           &              31.5            &        46.6           \\
CMT \cite{socher2013zero}                 &         58.8          &       61.4     &       35.8   &        37.0         &   48.3          \\
ReViSE$^a$        &     60.2        &    60.1       &    33.1      &       32.0    &       46.4          \\
ReViSE$^b$        &     64.4      &     63.2     &     36.0        &              37.4            &      50.3             \\ \hline \hline
\multicolumn{6}{|c|}{using both labeled and unlabeled training data}   \\ \hline \hline
DeViSE* \cite{frome2013devise}             &    65.6      &       58.9       &           35.4     &          31.3          &    47.8                \\
CMT* \cite{socher2013zero}       &    63.5    &    57.1    &     39.7     &              38.0          &   49.6            \\
ReViSE$^c$        &    66.8         &     63.6     &     39.4      &    37.5   &       51.8       \\
ReViSE       &    {\bf 74.2 }       &     {\bf 68.1 }   &    {\bf 47.6 }      &  {\bf 40.4  }    &        {\bf 57.6}      \\ \hline
\end{tabular}
}
\label{tbl:induc_zero_retr}
\vspace{-4mm}
\end{table}

\subsection{Zero-Shot Learning}
\label{subsec:zero_learn}
{
	Following the partitioning strategy of \cite{akata2015evaluation,xian2016latent}, we split $\mathsf{AwA}$ dataset into {\em 30/10/10} classes and $\mathsf{CUB}$ dataset into {\em 100/50/50} classes for {\em labeled training}/ {\em unlabeled training}/ {\em test} data. We adopt $\mathit{att}$ attributes as a textual description of each class. For zero-shot learning, not only the labels of images are unknown in the {\em unlabeled training} and {\em test} set, but classes are also disjoint across {\em labeled training}/ {\em unlabeled training}/ {\em test} splits. %Therefore, recognition and retrieval remains challenging without proper alignment of visual and textual information.

	To verify how unlabeled training data could benefit the learning of ReViSE, we provide four variants: ReViSE$^a$, ReViSE$^b$, ReViSE$^c$, and ReViSE. ReViSE$^a$ is when we only consider supervised objective. That is, $\alpha = 0$ in eq.~\eqref{eq:all}. ReViSE$^b$ is when we further take unsupervised objective in {\em labeled training data} into account; that is, only $\mathcal{L}_{reconstruct}$ and $\mathcal{L}_{MMD}$ are considered in $\mathcal{L}_{unsupervised}$ (see eq.~\eqref{eq:unsup}) for labeled training data. Next, for ReViSE$^c$, we consider {\em unlabeled training data} in $\mathcal{L}_{unsupervised}$ without unsupervised-data adaptation technique (setting $\beta = 0$). Last, ReViSE denotes our complete training architecture. 

	For completeness, we also consider the technique of unsupervised-data adaptation inference (see section~\ref{subsec:learning}) for DeViSE \cite{frome2013devise} and CMT \cite{socher2013zero}. In other words, we also evaluate how DeViSE and CMT benefit from the unlabeled training data. We adopt the same procedure as in eq.~\eqref{eq:test_adaptation} and report results as DeViSE* and CMT*, respectively.

	Similar to \cite{zhang2015zero,zhang2016zero1,zhang2016zero2}, the results and comparisons are reported using top-1 classification accuracy {\bf (top-1)} (see Table \ref{tbl:induc_zero_recog}) and mean average precision {\bf (mAP)} (see Table \ref{tbl:induc_zero_retr}) for recognition and retrieval tasks, respectively, on the unlabeled training and test images. To be more specific, we define the prediction score as \scalebox{0.9}{$\hat{y}_{i,c}^{(\cdot)} = \left \langle f'_v(\tilde{v}_{h,i}^{(\cdot)}), f'_t(t_{h,c}^{(\cdot)}) \right \rangle$} for a given image $v_i^{(\cdot)}$ and textual attributes $t_{c}^{(\cdot)}$ for class $c$. Results are provided after ranking $\hat{y}_{i,c}^{(\cdot)}$ on all unlabeled training or test classes. 

	Table \ref{tbl:induc_zero_recog} and \ref{tbl:induc_zero_retr} list the results for our zero-shot recognition and retrieval experiments. We first observe that {\bf NOT} all the methods benefit from using unlabeled training data during training. For example, in $\mathsf{AwA}$ dataset for test images $\mathbf{V_{te}}$, there is a $2.7\%$ retrieval deterioration from DeViSE to DeViSE* and a $3.1\%$ recognition deterioration from CMT to CMT*. On the other hand, our proposed method enjoys $2.6\%$ recognition improvement and $0.4\%$ retrieval improvement from ReViSE$^b$ to ReViSE$^c$. This shows that the learning method of our proposed architecture can actually benefit from unlabeled training data $\mathbf{V_{ut}}$ and $\mathbf{T_{ut}}$. 

	Next, we examine different variants in our proposed architecture. Comparing the average results from ReViSE$^a$ to ReViSE$^b$, we observe $3.4\%$ recognition improvement and $3.9\%$ retrieval improvement. This indicates that taking unsupervised objectives $\mathcal{L}_{reconstruct}$ and $\mathcal{L}_{MMD}$ into account results in learning better feature representations and thus yields a better recognition/ retrieval performance. Moreover, when unsupervised-data adaptation technique is introduced, we enjoy $5.0\%$ average recognition improvement and $5.8\%$ average retrieval improvement from ReViSE$^c$ to ReViSE. It is worth noting that the significant performance improvement for unlabeled training images $\mathbf{V_{ut}}$ further verifies that our unsupervised-data adaptation technique leads to a more accurate prediction on $\mathbf{V_{ut}}$.
}

\subsection{Transductive Zero-Shot Learning}
\label{subsec:new_zero_learn}
{

	In this subsection, we extend our experiments to a transductive setting, where test data are available during training. Therefore, the test data can now be regarded as the unlabeled training data ($\mathbf{V_{tr}} = \mathbf{V_{ut}}$ and $\mathbf{T_{tr}} = \mathbf{T_{ut}}$). To perform the experiments, as in Table \ref{tbl:dataset_stat}, we split $\mathsf{AwA}$ dataset into {\em 40/10} disjoint classes and $\mathsf{CUB}$ dataset into {\em 150/50} disjoint classes for {\em labeled training}/ {\em test} data.

	In order to evaluate different components in ReViSE, we further provide two variants: ReViSE$^{\dagger}$ and ReViSE$^{\dagger\dagger}$. ReViSE$^{\dagger}$ is when we consider no distributional matching between the codes across modalities ($\beta = 0$). ReViSE$^{\dagger\dagger}$ is when we further consider no contractive loss in our visual auto-encoder ($\beta = \gamma = 0$). Similar to previous subsection, we also consider DeViSE*, CMT*, and ReViSE$^{c}$ to evaluate the effect of our unsupervised-data adaptation inference.

\begin{table}[t!]
\centering
\caption{\footnotesize Transductive zero-shot recognition using top-1 classification accuracy (\%). }
\vspace{1mm}
\scalebox{0.72}
{
\begin{tabular}{|c||lll||lll||c|}
\hline
$\mathsf{Dataset}$ & \multicolumn{3}{c||}{$\mathsf{AwA}$}                                                      & \multicolumn{3}{c||}{$\mathsf{CUB}$}                                                     & average                          \\
                    $\textit{attributes}$    & \multicolumn{1}{c}{$\textit{att}$} & \multicolumn{1}{c}{$\textit{w2v}$} & \multicolumn{1}{c||}{$\textit{glo}$} & \multicolumn{1}{c}{$\textit{att}$} & \multicolumn{1}{c}{$\textit{w2v}$} & \multicolumn{1}{c||}{$\textit{glo}$} & top-1 acc. \\ \hline\hline
\multicolumn{8}{|c|}{test data not available during training}                                                                                                                                                                                     \\ \hline \hline
DeViSE \cite{frome2013devise}                  &           67.4               &           67.0             &        66.7                  &            40.8             &          28.8               &              25.6            &        49.3                         \\
CMT \cite{socher2013zero}                    &               67.6           &             69.5            &           68.0               &               42.4          &             29.6            &            25.7              &        50.5                         \\ \hline \hline
\multicolumn{8}{|c|}{test data available during training}                                                                                                                                                                                        \\ \hline \hline
DeViSE* \cite{frome2013devise}                &            90.7              &           84.8            &        88.0                  &           41.4              &            31.6             &              26.9            &       60.6                          \\
CMT* \cite{socher2013zero}                   &              89.4            &              87.8           &             81.8             &               43.1          &             31.8          &              28.9            &         60.5                       \\
ReViSE$^{\dagger\dagger}$             &            92.1              &             92.3            &         90.3      &       62.4          &       30.0            &         27.5       &           65.8                 \\
ReViSE$^{\dagger}$ &              92.8            &              92.6           &           91.7               &     62.7           &      31.8             &      28.9          &       66.8                     \\
ReViSE$^{c}$             &            73.0              &             67.0            &         73.4      &       53.7          &       26.4            &         28.2       &           53.6                \\
ReViSE                &           {\bf 93.4}               &           {\bf 93.5}      &            {\bf 92.2}       &         {\bf 65.4}         & {\bf 32.4}        &           {\bf 31.5}        &      {\bf 68.1}                 \\ \hline
\end{tabular}
}
\label{tbl:zero_recog}
\vspace{-4mm}
\end{table}

\vspace{0.1in}
\hspace{-5mm} {\bf Zero-Shot Recognition}:
Table \ref{tbl:zero_recog} reports top-1 classification accuracy. Observe that ReViSE clearly outperforms other state-of-the-art methods by a large margin. %For example, when we use $\mathsf{AwA}$ dataset with $\textit{att}$ attributes, we have $21.5\%$ performance improvement against LatEm. 
On average, we have at least $17\%$ gain compared to the methods without using unsupervised objective and $7.5\%$ gain compared to DeViSE* and CMT*. Note that all the methods work better on human annotated attributes ($\textit{att}$) than on unsupervised attributes ($\textit{w2v}$ and $\textit{glo}$) in $\mathsf{CUB}$ dataset. One possible reason is that for visually and semantically similar classes in a fine-grained dataset ($\mathsf{CUB}$), attributes obtained in an unsupervised way ($\textit{glo}$ word vectors) cannot fully differentiate between them. 
%However, when we use $\mathsf{CUB}$ dataset with $\textit{glo}$ attributes, our proposed method performs slightly worse than LatEm. One possible reason is that for visually and semantically similar classes in a fine-grained dataset ($\mathsf{CUB}$), attributes obtained in an unsupervised way ($\textit{glo}$ word vectors) cannot fully differentiate between them. This is also why all the methods work better on human annotated attributes ($\textit{att}$) than on unsupervised attributes ($\textit{w2v}$ and $\textit{glo}$) in $\mathsf{CUB}$ dataset. 
Nonetheless, for the more general concept dataset $\mathsf{AwA}$, using either supervised or unsupervised textual attributes, the performance does not differ by that much. 
For instance, our method achieves comparable performance using $\textit{att}$, $\textit{w2v}$, and $\textit{glo}$ ($93.4\%$, $93.5\%$, and $92.2\%$ top-1 classification accuracy) on $\mathsf{AwA}$ dataset.

The recognition performance for DeViSE* and CMT* ($60.6\%$ and $60.5\%$ on average) compared to DeViSE and CMT ($49.3\%$ and $50.5\%$ on average) further verifies that using unsupervised-data adaptation inference technique does benefit transductive zero-shot recognition. 
%As a recall, DeViSE* and CMT* are the variants of DeViSE and CMT when introducing unsupervised-data adaptation inference. 
Furthermore, all of the variants of ReViSE using unsupervised-data adaptation inference (ReViSE$^{\dagger\dagger}$, ReViSE$^\dagger$, and ReViSE itself) have noticeable improvement over DeViSE* and CMT*. 
This shows that the proposed model succeeds in leveraging unsupervised information in test data for constructing more effective cross-modal embeddings.

Next, we evaluate the effects of different components designed in our architecture. First of all, we compare the results between ReViSE$^\dagger$ (set $\beta = 0$) and ReViSE. The performance gain ($66.8\%$ to $68.1\%$ on average) indicates that minimizing MMD distance between visual and textual codes enables our model to learn more robust visual-semantic embeddings. In other words, we can better associate cross-modal information when we match the distributions across visual and textual domains (please refer to Supplementary for the study of MMD distance). Second, we observe that, without contractive loss, performance slightly drop from $66.8\%$ (ReViSE$^\dagger$) to $65.8\%$ (ReViSE$^{\dagger\dagger}$). This is not surprising since the contractive auto-encoder aims at learning less varying features/codes with similar visual input, and therefore we can expect to learn more robust visual codes. 
Finally, similar to the observations found in comparing DeViSE/CMT to DeViSE*/CMT*, the {\em unsupervised-data adaptation inference} in ReViSE substantially improves the average top-1 classification 
accuracy from $53.6\%$ (ReViSE$^c$) to $68.1\%$ (ReViSE).
Please see Supplementary material for more detailed comparisons to the following non-deep-embeddings methods: SOC \cite{palatucci2009zero}, ConSE \cite{norouzi2013zero}, SSE \cite{zhang2015zero}, SJE \cite{akata2015evaluation}, ESZSL \cite{romera2015embarrassingly}, JLSE \cite{zhang2016zero1}, LatEm \cite{xian2016latent}, Sync \cite{changpinyo2016synthesized}, MTE \cite{bucher2016improving}, TMV \cite{fu2015transductive}, and SMS \cite{guo2016transductive}.

\vspace{0.1in}
\hspace{-5mm} {\bf Zero-Shot Retrieval}:
In Table \ref{tbl:zero_retr}, we report zero-shot retrieval results by measuring the retrieval performance by mean average precision (mAP). On average, methods that leverage unsupervised information yield better performance compared to the methods using no unsupervised objective. However, in few cases, the performance drops when we take unsupervised information into account. For example, on $\mathsf{CUB}$ dataset, DeViSE* performs unfavorably compared to DeViSE when $\textit{w2v}$ and $\textit{glo}$ word embeddings are used as textual attributes. %We note that DeViSE* is the variant of DeViSE with unsupervised-data adaptation inference. Therefore, if we have a low retrieval performance (i.e., $24.5\%$ for DeViSE in $\mathsf{CUB}$ with $\textit{glo}$ attributes) originally, then performing the unsupervised-data adaptation inference might not be desirable. 

Overall, our method does help improve zero-shot retrieval by at least $14.1\%$ compared to CMT*/DeViSE* and $21.5\%$ compared to CMT/DeViSE. It clearly demonstrates the effectiveness of leveraging unsupervised information for improving zero-shot retrieval (please see Supplementary for the plot of precision-recall curves).

In addition to quantitative results, we also provide qualitative results of ReViSE. Fig. \ref{fig:retr_image} is the image retrieval experiments
for classes {\em Chestnut\_sided\_Warbler} and {\em White\_eyed\_Vireo}. Given a class embedding, the nearest image neighbors are retrieved based on the cosine similarity between transformed visual and textual features. We consider two conditions: images from the same class and images from all test classes. In {\em Chestnut\_sided\_Warbler}, most of the images ($71.7\%$) are correctly classified, and we also observe that three nearest image neighbors are also in {\em Chestnut\_sided\_Warbler}. On the other hand, only $43.3\%$ images are correctly classified in {\em White\_eyed\_Vireo}, and two of the three nearest image neighbors are form wrong class {\em Wilson\_Warbler}.

\begin{figure}[t!]
\includegraphics[width=0.47\textwidth]{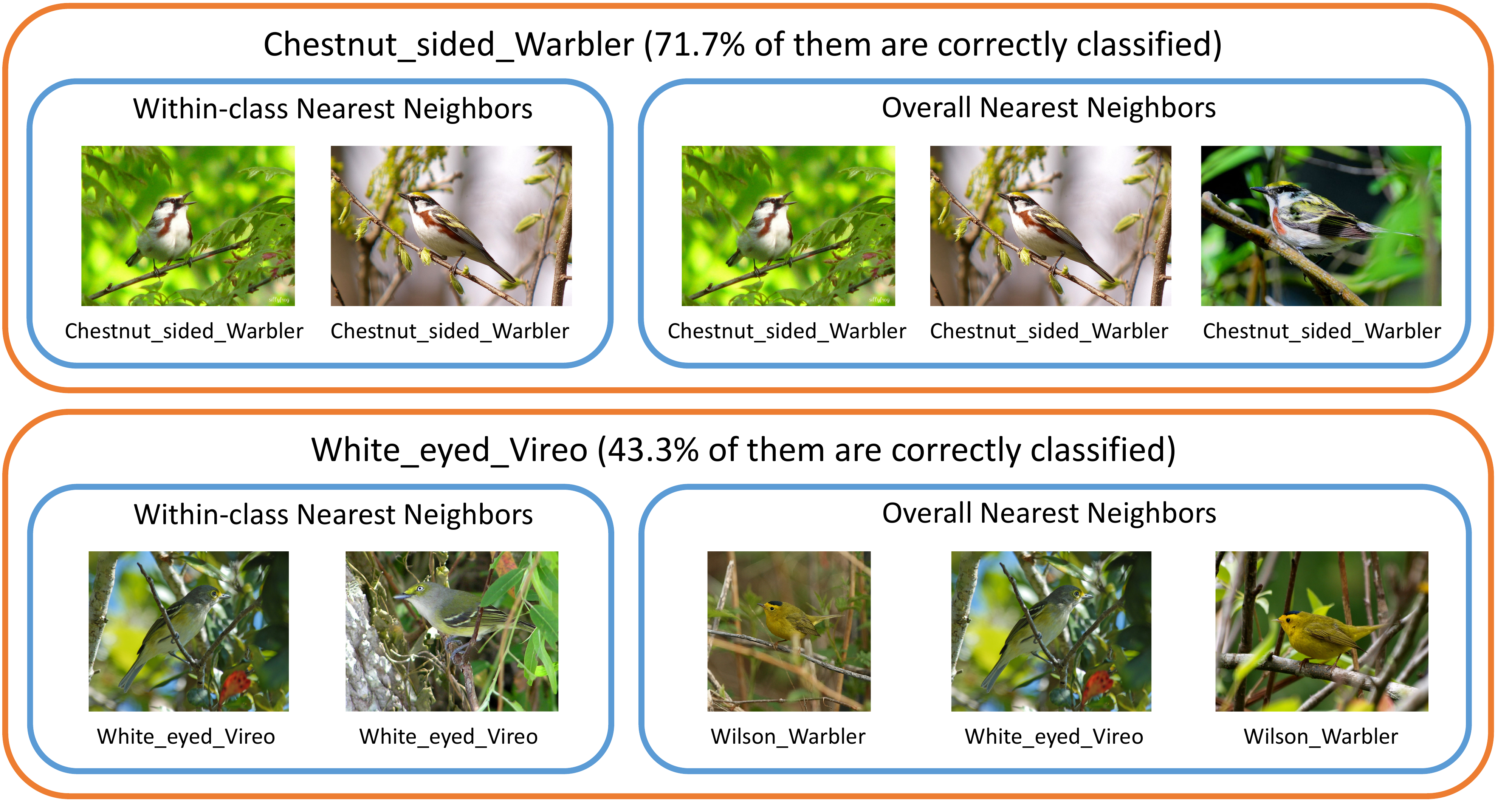}
\caption{\footnotesize Images-retrieval experiments for $\mathsf{CUB}$ with $\textit{att}$ attributes.}
\label{fig:retr_image}
\vspace{-0.1in}
\end{figure}

\begin{table}[t!]
\centering
\caption{\footnotesize Transductive zero-shot retrieval using mean Average Precision (mAP) (\%).}
\vspace{1mm}
\scalebox{0.72}
{
\begin{tabular}{|c||lll||lll||c|}
\hline
$\mathsf{Dataset}$ & \multicolumn{3}{c||}{$\mathsf{AwA}$}                                                      & \multicolumn{3}{c||}{$\mathsf{CUB}$}                                                     & average                          \\
                    $\textit{attributes}$    & \multicolumn{1}{c}{$\textit{att}$} & \multicolumn{1}{c}{$\textit{w2v}$} & \multicolumn{1}{c||}{$\textit{glo}$} & \multicolumn{1}{c}{$\textit{att}$} & \multicolumn{1}{c}{$\textit{w2v}$} & \multicolumn{1}{c||}{$\textit{glo}$} & mAP \\ \hline\hline
\multicolumn{8}{|c|}{test data not available during training}                                                                                                                                                                                     \\ \hline \hline
DeViSE \cite{frome2013devise}                  &              67.5            &             67.6           &         66.2                 &            31.9             &           26.6              &           24.5               &        47.4              \\
CMT \cite{socher2013zero}                    &              66.3            &             70.6            &         69.5                 &             39.3            &             25.2            &           21.9              &     48.8                       \\ \hline \hline
\multicolumn{8}{|c|}{test data not available during training}                                                                                                                                                                                        \\ \hline \hline
DeViSE* \cite{frome2013devise}                &              82.3            &              78.0          &        84.4                  &             36.9            &            25.8             &             21.3             &       54.8               \\
CMT* \cite{socher2013zero}                   &              85.8            &             77.3            &       73.0                   &              44.1           &            28.9             &             28.1            &     56.2                     \\
ReViSE$^{\dagger\dagger}$&            96.7              &            96.8             &         95.1          &     60.7             &       29.4           &        27.2        &            67.7            \\
ReViSE$^{\dagger}$&              97.2            &               96.9          &          96.4                &    62.0           &       29.8            &        28.2       &        68.4          \\
ReViSE$^{c}$             &            73.0              &             67.0            &         73.4      &       53.7          &       26.4            &         28.2       &           53.6                \\
ReViSE                &          {\bf 97.4}                &          {\bf 97.4}               &         {\bf 96.7}            &      {\bf 68.9}           &       {\bf 30.5}      &        {\bf 30.9}          &          {\bf 70.3}           \\ \hline
\end{tabular}
}
\vspace{-4mm}
\label{tbl:zero_retr}
\end{table}

\vspace{0.1in}
\hspace{-5mm} {\bf Availability of Unlabeled Test Images}:
We next evaluate the performance of our method w.r.t. to the availability of test images for unsupervised objective (see Fig. \ref{fig:test_unsup}) on $\mathsf{CUB}$ dataset with $\textit{att}$ attributes. We alter the fraction $p$ of unlabeled test images used in the training stage from $0\%$ to $100\%$ by a step size of $10\%$. That is, in eq. \eqref{eq:unsup}, only $p$ portion (randomly chosen) of test images contributes to $\mathcal{L}_{unsupervised}$. 
%However, all unseen images are revealed in test time. 
Fig. \ref{fig:test_unsup} clearly indicates the performance increases when $p$ increases. That is, with more unsupervised information (test images) available, our model can better associate the supervised and unsupervised data.
%It is noteworthy that when no unlabeled images available ($p = 0$) during training time, only images from training classes contributes to $\mathcal{L}_{recontruct}$ and $\mathcal{L}_{MMD}$ in eq. \eqref{eq:unsup}. 
Another interesting observation is that with only $40\%$ test images available, ReViSE achieves favorable performance on both transductive zero-shot recognition and retrieval. 

\hspace{-5mm} {\bf Expand the test-time search space:} Note that most of the methods \cite{frome2013devise,socher2013zero,norouzi2013zero,zhang2015zero,akata2015evaluation,zhang2016zero1,xian2016latent,changpinyo2016synthesized,fu2015transductive} consider that, at test time, queries come from only test classes. For $\mathsf{AwA}$ dataset with $\textit{att}$ attributes, we expand the test-time search space to all training and test classes and perform transductive zero-shot recognition for DeViSE*, CMT*, and ReViSE. We discover severe performance drops from $90.7\%$, $89.4\%$, and $93.4\%$ to $47.4\%$, $45.8\%$, and $42.5\%$. Similar results can also be observed in other non-deep-embeddings methods. Although challenging, it remains interesting to consider this generalized zero-/few-shot learning setting in our future work.

\begin{figure}[t!]
\centering
\includegraphics[width=0.43\textwidth]{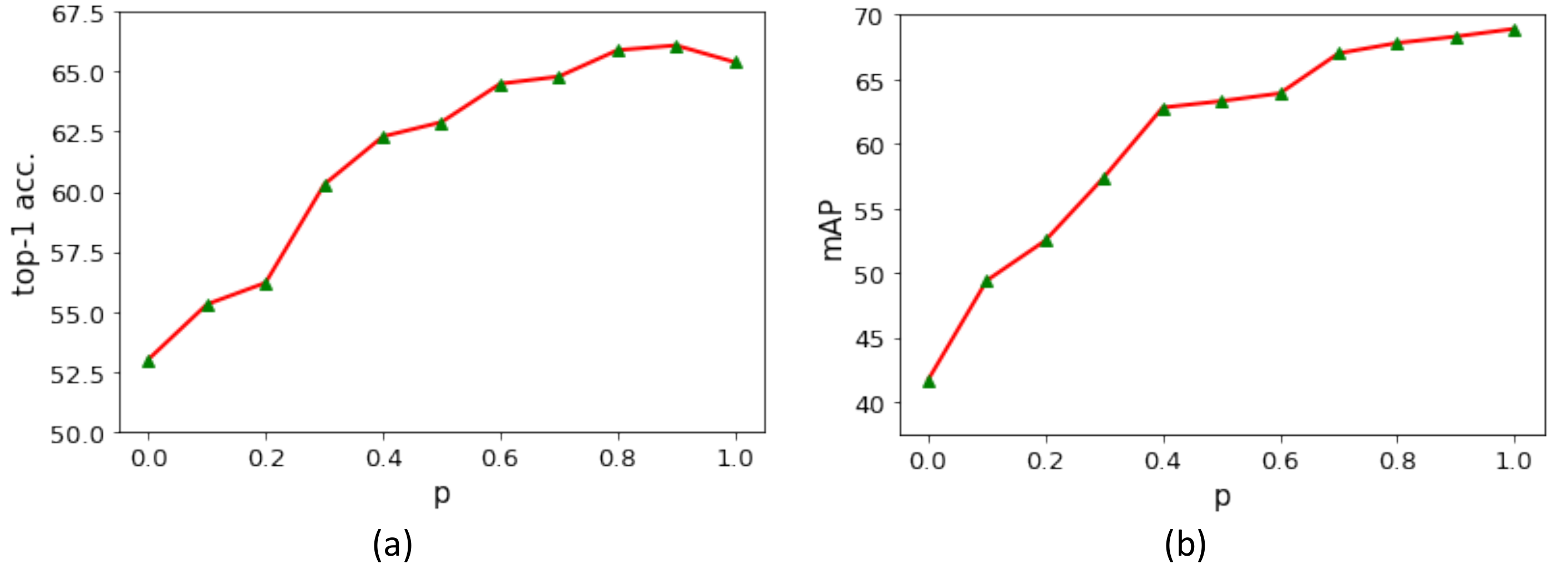}
\caption{\footnotesize Fraction $p$ of test images used for training ReViSE on transductive (a) zero-shot recognition (b) zero-shot retrieval for $\mathsf{CUB}$ dataset with $\textit{att}$ attributes.}
\label{fig:test_unsup}
\vspace{-3mm}
\end{figure}

}

\begin{figure*}[t!]
\vspace{-3mm}
\centering
\includegraphics[width=0.8\textwidth]{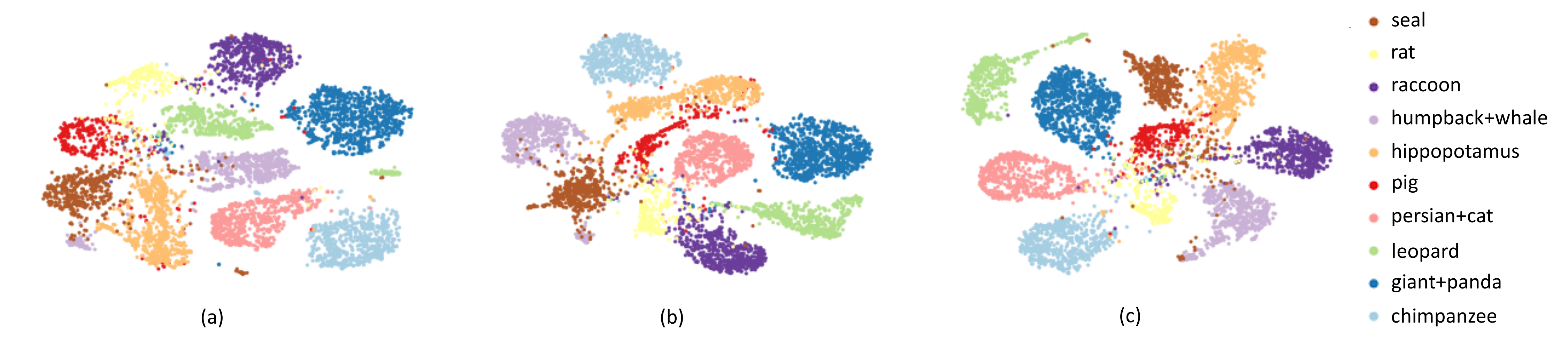}
\caption{\footnotesize (a) Original CNN features (b) Reconstructed features (c) Visual codes for $\mathsf{AwA}$ dataset in ReViSE under transductive zero-shot setting. We use $\textit{glo}$ as our textual attributes for classes. Different colors denote different classes. Best viewed in colors.}
\label{fig:orig_recon_code}
\vspace{-4mm}
\end{figure*}

\begin{figure*}[t!]
\centering
\includegraphics[width=0.8\textwidth]{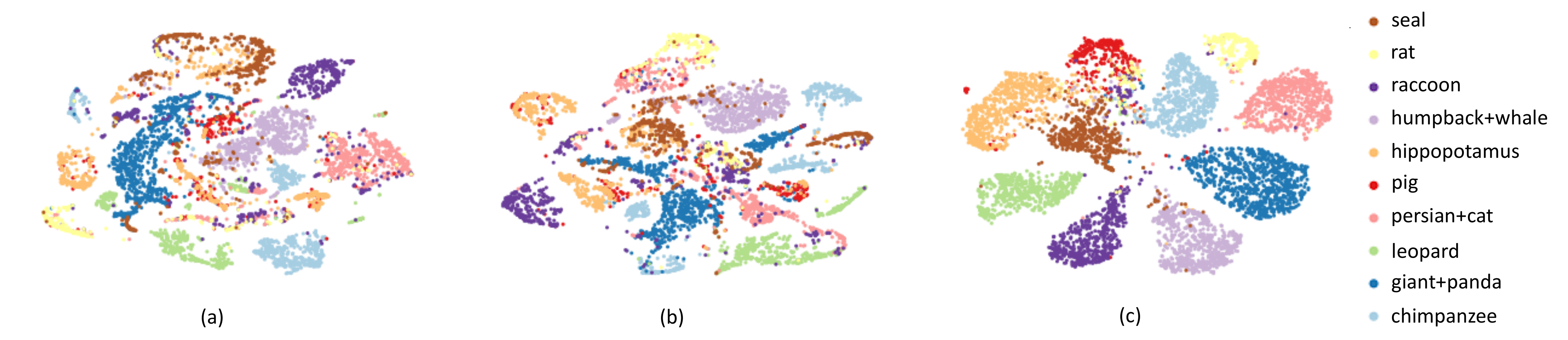}
\caption{\footnotesize Output features of : (a) DeViSE* (b) CMT* (c) ReViSE. $\textit{glo}$ attributes are used on $\mathsf{AwA}$ dataset under transductive zero-shot setting. Different colors denote different classes. Best view in colors.}
\label{fig:devise_cmt_ours}
\vspace{-3mm}
\end{figure*}

\subsection{From Zero to Few-Shot Learning}
\label{subsec:few_shot}
{

In this subsection, we extend our experiments from transductive zero-shot to transductive few-shot learning. 
Compared to zero-shot learning, few-shot learning allows us to have a few labeled images in our test classes. Here, 3 images are randomly chosen to be labeled per test category. We use the same performance comparison metrics as in Sec. \ref{subsec:zero_learn} to report the results.

\vspace{0.1in}
\hspace{-5mm} {\bf Transductive Few-Shot Recognition and Retrieval}:
Tables \ref{tbl:few_recog} and \ref{tbl:few_retriev} list the results of transductive few-shot recognition and retrieval tasks. Generally speaking, ReViSE achieves the best performance compared to its variants and other methods. Moreover, as expected, when we compare the results with transductive zero-shot recognition (Table \ref{tbl:zero_recog}) and retrieval (Table \ref{tbl:zero_retr}), every methods perform better when few (i.e., $3$) labeled images are observed in the test classes. For example, for $\mathsf{CUB}$ dataset with $\textit{w2v}$ attributes, there is a $22.5\%$ recognition improvement for CMT* and a $32.3\%$ retrieval improvement for ReViSE. 

We also observe that the performance gap between our proposed ReViSE and other methods is reduced compared to transductive zero-shot learning. For instance, in average retrieval performance, ReViSE has $15.5\%$ mAP improvement over DeViSE* under zero-shot experiments, while only $9.3\%$ improvement under few-shot experiments.

\begin{table}[t!]
\centering
\caption{\footnotesize Few-shot recognition comparison using top-1 classification accuracy (\%). For each test class, 3 images are randomly labeled, while the rest are unlabeled.}
\vspace{1mm}
\scalebox{0.72}
{
\begin{tabular}{|c||lll||lll||c|}
\hline
$\mathsf{Dataset}$ & \multicolumn{3}{c||}{$\mathsf{AwA}$}                                                      & \multicolumn{3}{c||}{$\mathsf{CUB}$}                                                     & average                          \\
                    $\textit{attributes}$    & \multicolumn{1}{c}{$\textit{att}$} & \multicolumn{1}{c}{$\textit{w2v}$} & \multicolumn{1}{c||}{$\textit{glo}$} & \multicolumn{1}{c}{$\textit{att}$} & \multicolumn{1}{c}{$\textit{w2v}$} & \multicolumn{1}{c||}{$\textit{glo}$} & top-1 acc. \\  \hline \hline
\multicolumn{8}{|c|}{test data not available during training}                                                                                                                                                                                        \\ \hline \hline
DeViSE \cite{frome2013devise}     &   80.9     &   75.3 &  79.4  & 54.0  &      45.7  &  46.0   &  63.6 \\
CMT \cite{socher2013zero}    &  85.1  &    83.4    &  84.3   &    56.7     &    53.4   &    52.0   & 69.2  \\ \hline \hline
\multicolumn{8}{|c|}{test data available during training}                                                                                                                                                                                        \\ \hline \hline
DeViSE* \cite{frome2013devise}     &    92.6      &      91.1  &   91.3     &   57.5     &      50.7   &   52.9       &    72.7   \\
CMT* \cite{socher2013zero}    &  90.6  &    90.2    &  91.1   &    62.5     &    54.3   &    55.4   &  74.0    \\
ReViSE$^{\dagger\dagger}$ &  93.3    &    93.3     &   93.1   &   66.9      &    57.6    &   59.0    &   77.2  \\
ReViSE$^{\dagger}$ &   93.3  &  93.8  &     93.5       &  67.7  &   59.6    &     60.0    &   78.0  \\
ReViSE$^{c}$ &   87.8 &  88.7 &    90.2     &  61.1 &   55.3  &  55.0   &  73.0 \\
ReViSE     &    {\bf 94.2}    &    {\bf 94.1}   &  {\bf 94.4}  &   {\bf 68.4}   &  {\bf 59.9} &  {\bf 61.7}    &    {\bf 78.8}    \\ \hline
\end{tabular}
}
\vspace{-4mm}
\label{tbl:few_recog}
\end{table}

}

\subsection{t-SNE Visualization}
\label{subsec:tsne}
{

Figure~\ref{fig:orig_recon_code} further shows the t-SNE \cite{maaten2008visualizing} visualization for the original CNN features, the reconstructed visual features $r_v(\tilde{v}^{(te)})$, and the visual codes $\tilde{v}_h^{(te)}$ on $\mathsf{AwA}$ dataset with $\textit{glo}$ attributes under transductive zero-shot setting. 
First of all, observe that both the reconstructed features and the visual codes have more separate clusters over different classes, which suggest ReViSE has learned useful representations. Another interesting observation is that the affinities between classes might change after learning visual codes. For example, ``leopard'' images (green dots) are near ``humpback whale'' images (light purple dots) in the original CNN feature space. However, in the visual code space, leopard images are far from humpback whale images. One possible explanation is that we know leopard is semantically distinct from humpback whale, and thus their semantic attributes must also be very different. This leads to different image clusters in our designed framework.

Next, we provide the t-SNE visualization on the output visual test scores $f_v(\mathbf{V_{te}})$ for DeViSE*, CMT*, and ReViSE in Fig. \ref{fig:devise_cmt_ours}. Clearly, ReViSE can better separate instances from different classes. 

%In Fig. \ref{fig:devise_cmt_ours}, we provide the t-SNE \cite{maaten2008visualizing} visualization on the output visual scores $f_v(\mathbf{V_{te}})$ for DeViSE*, CMT*, and our proposed method. Clearly, our proposed method can better separate instances from different classes and thus be consider as a more robust visual-semantic embedding.

}
\begin{table}[t!]
\centering
\caption{\footnotesize Few-shot retrieval comparison using mean Average Precision (mAP) (\%). For each test class, 3 images are randomly labeled, while the rest are unlabeled. }
\vspace{1mm}
\scalebox{0.72}
{
\begin{tabular}{|c||lll||lll||c|}
\hline
$\mathsf{Dataset}$ & \multicolumn{3}{c||}{$\mathsf{AwA}$}                                                      & \multicolumn{3}{c||}{$\mathsf{CUB}$}                                                     & average                          \\
                    $\textit{attributes}$    & \multicolumn{1}{c}{$\textit{att}$} & \multicolumn{1}{c}{$\textit{w2v}$} & \multicolumn{1}{c||}{$\textit{glo}$} & \multicolumn{1}{c}{$\textit{att}$} & \multicolumn{1}{c}{$\textit{w2v}$} & \multicolumn{1}{c||}{$\textit{glo}$} & mAP \\  \hline \hline
\multicolumn{8}{|c|}{test data not available during training}                                                                                                                                                                                        \\ \hline \hline
DeViSE \cite{frome2013devise}     &   85.0  &   79.3  &  84.9  &   46.4    &    42.6   & 42.9       &  63.5  \\
CMT \cite{socher2013zero}    &  88.4  &    88.2    &  89.2   &    58.5     &    54.0   &    52.7   &  71.8   \\ \hline \hline
\multicolumn{8}{|c|}{test data available during training}                                                                                                                                                                                        \\ \hline \hline
DeViSE* \cite{frome2013devise}      &      96.7   &     95.5      &   95.8     &47.5     &   49.2   &   51.6    & 72.7   \\
CMT* \cite{socher2013zero}    &   95.3   &    94.8       &  95.8  &   60.0  &  54.7    &    56.4  &  76.2   \\
ReViSE$^{\dagger\dagger}$ &  97.2   &   97.1     &    97.1   &   71.2   &   59.4   &   61.4  &    80.6     \\
ReViSE$^{\dagger}$ &   97.3    &   97.5    &   97.4      &   72.5    &   61.4   &  62.5    &   81.4  \\
ReViSE$^{c}$ &  92.3  & 93.0  &  94.6    &  60.8 &  55.0 &  57.1   &  75.5 \\
ReViSE   &     {\bf 97.8}   &  {\bf 97.7}    &  {\bf 97.8}  &    {\bf 72.9}  &    {\bf 62.8}  &   {\bf 63.0}    &    {\bf 82.0} \\ \hline
\end{tabular}
}
\vspace{-4mm}
\label{tbl:few_retriev}
\end{table}

}

\vspace{-1mm}
\section{Conclusion}
{
In this paper, we showed how we can augment a typical supervised formulation with unsupervised techniques for learning joint embeddings of visual and textual data. We empirically evaluate our proposed method on both general and fine-grained image classification datasets, with comparisons against the state-of-the-art methods in zero-shot and few-shot recognition and retrieval tasks, from inductive to transductive setting. 
In all the experiments, our method consistently outperforms other methods, substantially 
improving performance in some cases. We believe that this work sheds light on the advantages of combining supervised and unsupervised learning techniques, and makes a step towards learning 
more useful representations from multi-modal data.
}

{\small
\bibliographystyle{ieee}
\bibliography{egbib}
}
\newpage
\section{Network Design}
{
	Fig. \ref{fig:network} provides an easy-to-understand design of ReViSE. In all of our experiments, GoogLeNet is pre-trained on ImageNet \cite{deng} images. Without fine-tuning, we directly extract the top layer activations (1024-dim) as our input image features followed by a common $log(1+v)$ pre-processing step. For the textual attributes, we pre-process them through a standard $l_2$ normalization.

	In ReViSE, we set $\alpha = 1.0$ in eq. (11), so that we place equal importance on supervised and unsupervised objectives. For the visual auto-encoder, we fix the parameter of the contraction strength $\gamma = 0.1$ in eq. (2). In the following, we omit the bias term in each layer for simplicity. The encoding of visual features is parameterized by a two-hidden layer fully-connected neural network with architecture $d_{v1}-d_{v2}-d_{c}$, where $d_{v1} = 1024$ is the input dimension of the visual features, $d_{v2} = 500$ is the intermediate layer, and $d_c$ denotes the dimension of the visual codes $\tilde{v}_h$. To encode textual attributes, we consider a single-hidden layer neural network $d_{t1}-d_{c}$, where $d_{t1}$ is the input dimension of the textual attributes. We choose $d_c = 100$ when $d_{t1} > 100$ and $d_c = 75$ when $d_{t1} < 100$. Furthermore, we do not tie the weights to be learned between the decoding and encoding parts. Parameters for associating distributions of visual and textual codes (MMD Loss) in eqs. (5) (12), and (6) are set as $\beta = \{0.1, 1.0\}$ (chosen by cross-validation) and $\kappa = 32.0$. For the remaining part of our model, we set the architecture of visual and textual code mapping as a single-hidden layer fully-connected neural network with dimension $d_c-50$. We also adopt a dropout of $0.7$. 

	During the first 100 iterations of training, we set $\lambda = 0$ so that no unsupervised-data adaptation is used while still updating $\hat{I}_{i,c}^{(ut)}$. Note that $\hat{I}_{i,c}^{(ut)}$ are the {\em inferred labels} for unsupervised data, and {\em not random} at each iteration. Beginning with the 101th iteration, we set $\lambda = \{0.1,1.0\}$ (chosen by cross-validation), and the model typically converges within 2000 to 5000 iterations.

	We implement ReViSE in TensorFlow \cite{tensor}. We use Adam \cite{adam} for optimization with minibatches of size $1024$. We choose $tanh$ for all of our activation functions.

	\begin{figure}[t!]
	\includegraphics[width=0.47\textwidth]{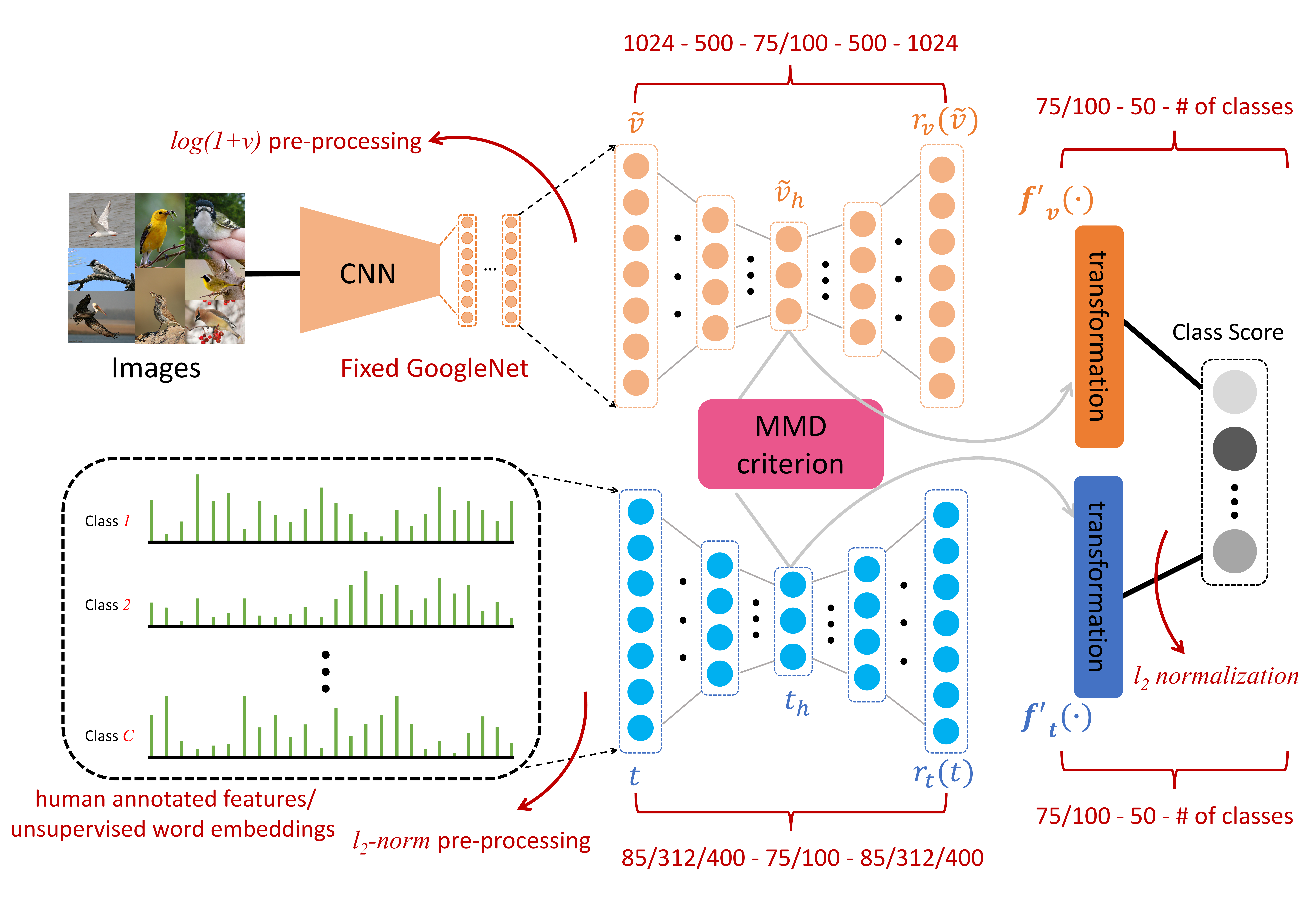}
	\caption{Our designed architecture.}
	\label{fig:network}
	\vspace{-0.1in}
	\end{figure}
}
	\begin{figure*}[t!]
	\centering
	\includegraphics[width=0.9\textwidth]{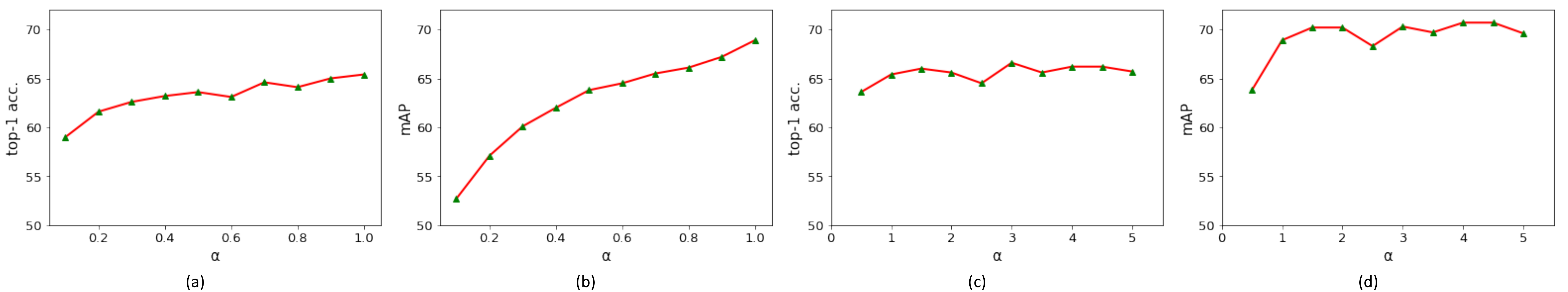}
	\caption{\footnotesize Varying $\alpha$ in two scales: $0.1$ to $1.0$ and $0.5$ to $5.0$. (a),(c) display  plots for transductive zero-shot recognition and (b),(d) display plots for transductive zero-shot retrieval. $\mathsf{CUB}$ dataset with $\textit{att}$ attributes are used in the experiments.}
	\label{fig:alpha}
	\vspace{-3mm}
	\end{figure*}
\section{Parameters Choice}
{
	We have four parameters in our architecture: $\alpha, \beta, \gamma$, and $\kappa$. We fix $\alpha = 1.0$, $\gamma = 0.1$, $\kappa = 32.0$ for all the experiments. Then we set $\lambda = 0.0$ (no unsupervised-data adaptation inference), and perform cross-validation on the splitting set as suggested by [3,46] to determine $\beta$ from $\{0.1, 1.0\}$. Next, with chosen $\beta$, we perform cross-validation to choose $\lambda$ from $\{0.1, 1.0\}$. Table \ref{tbl:beta_lambda} lists the statistics of $\beta$ and $\lambda$.
	\begin{table}[t!]
	\centering
	\caption{\footnotesize Value of $\beta$ and $\lambda$.}
	\vspace{1mm}
	\scalebox{0.8}
	{
	\begin{tabular}{|c||lll||lll|}
	\hline
	$\mathsf{Dataset}$ & \multicolumn{3}{c||}{$\mathsf{AwA}$}	                                                      & \multicolumn{3}{c|}{$\mathsf{CUB}$}	                                                 \\
	                    $\textit{attributes}$    & \multicolumn{1}	{c}{$\textit{att}$} & \multicolumn{1}{c}{$	\textit{w2v}$} & \multicolumn{1}{c||}{$\textit{glo}$} & \multicolumn{1}{c}{$\textit{att}$} & \multicolumn{1}{c}{$\textit{w2v}$} & \multicolumn{1}{c|}{$\textit{glo}$}  \\ \hline\hline
	$\beta$                     & 0.1                     & 1.0                   & 1.0                     & 1.0                   & 1.0                   & 1.0                     \\
	$\lambda$                  & 1.0                     & 1.0                & 1.0              & 1.0                 & 	0.1                 & 0.1                  \\ \hline
	\end{tabular}
	}
	\label{tbl:beta_lambda}
	\vspace{-3mm}
	\end{table}

	Next, we study the power of unsupervised information. We now take $\mathsf{CUB}$ dataset with $\textit{att}$ attributes to test the advantage of using unsupervised information, which can be viewed as tuning the parameter $\alpha$ for the unsupervised objective in eq. (11). Originally, $\alpha$ was set to $1.0$, which equally weights the contribution of supervised and unsupervised loss. We now alter $\alpha$ as follows: $0.1$ to $1.0$ by step size of $0.1$ and $0.5$ to $5.0$ by step size of $0.5$. The results are shown in Fig. \ref{fig:alpha}. We observe that when $\alpha$ increases from $0.1$ to $1.0$, the performance increases; however, when $\alpha$ increase from $1.0$ to $5.0$, the performance stays relatively unchanged. Empirically, we find that ReViSE does not perform better when $\alpha > 1.0$, which is expected, since we should not view unsupervised information more important than supervised information.
}
	\begin{figure}[t!]
	\centering
	\includegraphics[width=0.36\textwidth]{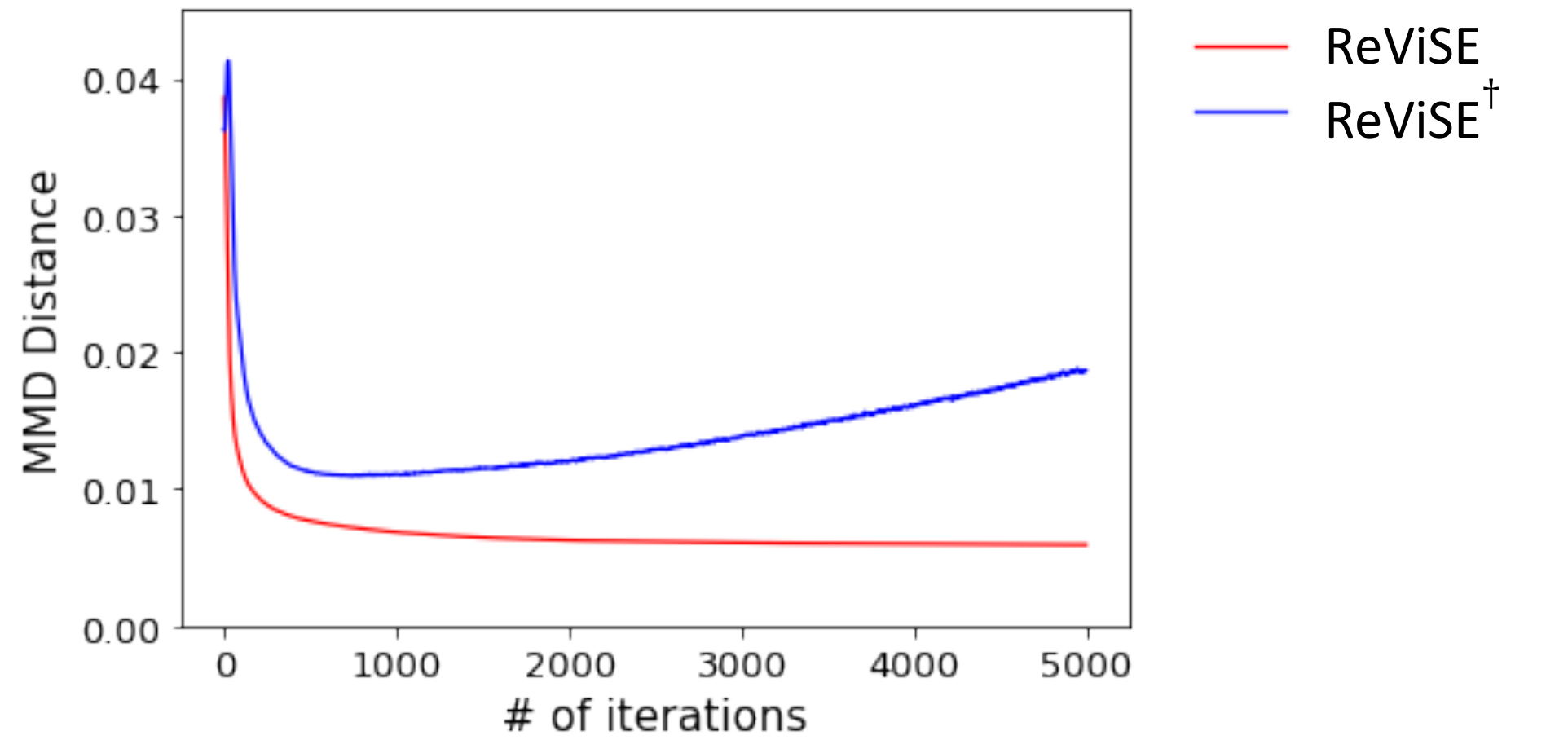}
	\caption{\footnotesize MMD distance w.r.t. \# of iterations for our method with and without $\mathcal{L}_{MMD}$. The experiment is conducted on $\mathsf{CUB}$ dataset with $\textit{att}$ attributes under transdutive zero-shot setting.}
	\label{fig:mmd_loss}
	\vspace{-2mm}
	\end{figure}

\section{Precision-Recall Curve}
{
	Fig. \ref{fig:CUB_mAP} is the precision-recall curve for zero-shot retrieval results on $\mathsf{CUB}$ dataset with $\textit{att}$ attributes.

	\begin{figure}[t!]
	\includegraphics[width=0.47\textwidth]{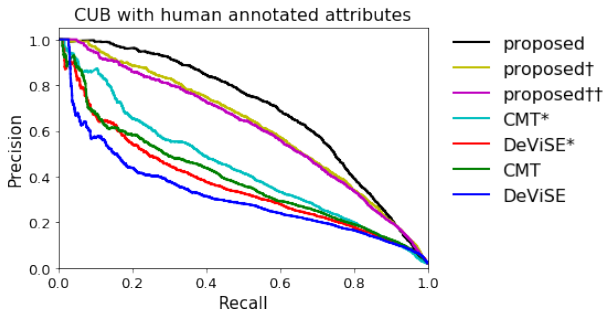}
	\caption{Precision-recall curve comparison for zero-shot retrieval on $\mathsf{CUB}$ with human annotated attributes as textual attributes for classes. Best viewed in color.}
	\label{fig:CUB_mAP}
	\end{figure}
}

\section{MMD Distance}
{
	MMD distance in eq. (5) can be viewed as the {\em distribution measurement} [13] between visual and textual code. For $\mathsf{CUB}$ dataset with $\textit{att}$ attributes under transductive zero-shot experiment, we calculate the MMD distance (on the test codes) in our method with (ReViSE) and without (ReViSE$^\dagger$) $\mathcal{L}_{MMD}$. The results of MMD distance w.r.t. the number of iterations are shown in Fig. \ref{fig:mmd_loss}. We clearly observe that the red curve (ReViSE) has consistently lower value than the blue curve (ReViSE$^\dagger$). Moreover, based on the previous results, ReViSE always performs better than ReViSE$^\dagger$. Hence aligning the distributions across visual and textual codes can better associate cross-modal information and thus lead to more robust visual-semantic embeddings.

}
\section{Remarks on Contractive Loss}
{
	We find that adding contractive loss to textual auto-encoder doesn't provide much benefit. One possible reason may be the limited number of textual features ($200$ for $\mathsf{CUB}$). On the other hand, the number of visual features is large ($11,786$ for $\mathsf{CUB}$).
}

\section{Comparing with recent state-of-the-art methods}
{
	In our main paper, we focus on comparing with deep-embeddings methods. In Table \ref{tbl:zero_recog}, we compare other methods for inductive and transductive zero-shot learning. Note that SMS$_{ESZSL}$ adopts ESZSL for its initialization.

	\begin{table}[t!]
	\centering
	
	\caption{\footnotesize Inductive and transductive zero-shot recognition using top-1 classification accuracy (\%). }
	\vspace{2mm}
	\scalebox{0.73}
	{
	\begin{tabular}{|c||lll||lll||c|}
	\hline
	$\mathsf{Dataset}$ & \multicolumn{3}{c||}{$\mathsf{AwA}$}                                                      & \multicolumn{3}{c||}{$\mathsf{CUB}$}                                                     & average                          \\
	                    $\textit{attributes}$    & \multicolumn{1}{c}{$\textit{att}$} & \multicolumn{1}{c}{$\textit{w2v}$} & \multicolumn{1}{c||}{$\textit{glo}$} & \multicolumn{1}{c}{$\textit{att}$} & \multicolumn{1}{c}{$\textit{w2v}$} & \multicolumn{1}{c||}{$\textit{glo}$} & top-1 acc. \\ \hline\hline
	\multicolumn{8}{|c|}{test data not available during training}                                                                                                                                                                                     \\ \hline \hline
	SOC [30]                  &           58.6               &          50.8           &      68.0           &     34.7         &       30.9         &          30.6         &      45.6                   \\
	ConSE [29]                    &   59.0        &       53.2        &    49.8              &       33.6        &        28.8           &       30.8             &     42.5                       \\ 
	SSE [49]                  &         63.8           &        58.6            &   65.8              &    31.8             &      27.9           &       25.4           &       45.6            \\
	SJE [2]                    &   66.7     &             52.1      &    58.8          &         50.1        &             28.4     &       24.2          &     46.7                 \\ 
	ESZSL [37]                  &     76.8           &    62.2          &        67.7              &      50.3            &          33.4        &        34.1           &        54.1                    \\
	JLSE [50]                    &         71.8         &             64.0       &       68.0             &        33.7       &             28.0        &        27.1          &        48.8                   \\
	LatEm [48]                  &       71.9            &     61.1       &    62.9              &      45.5            &          31.8        &           32.5          &        51.0            \\
	Sync [7]                    &   72.9       &      62.0     &     67.0        &      48.7         &       31.2          &            32.8        &    52.4           \\ 
	MTE [6]                    &          77.3        &             -       &      -          &         43.3         &             -      &      -           &      -                   \\ 
	DeViSE [9]                  &           67.4               &           67.0             &        66.7                  &            40.8             &          28.8               &              25.6            &        49.3                         \\
	CMT [41]                    &               67.6           &             69.5            &           68.0               &               42.4          &             29.6            &            25.7              &        50.5                         \\ \hline \hline
	\multicolumn{8}{|c|}{test data available during training}                                                                                                                                                                                        \\ \hline \hline
	TMV [10]                &         89.0          &           69.0       &    88.7                 &      51.2             &       32.5          &    {\bf 38.9}           &   61.6              \\
	SMS$_{ESZSL}$ [12]                &     89.6           &           78.0       &     82.9                &     52.3          &     {\bf 34.7}       &    32.3       &    61.6           \\
	DeViSE* [9]                &            90.7              &           84.8            &        88.0                  &           41.4              &            31.6             &              26.9            &       60.6                          \\
	CMT* [41]                  &              89.4            &              87.8           &             81.8             &               43.1          &             31.8          &              28.9            &         60.5                       \\
	ReViSE$^{\dagger\dagger}$             &            92.1              &             92.3            &         90.3      &       62.4          &       30.0            &         27.5       &           65.8                 \\
	ReViSE$^{\dagger}$ &              92.8            &              92.6           &           91.7               &     62.7           &      31.8             &      28.9          &       66.8                     \\
	ReViSE$^{c}$             &            73.0              &             67.0            &         73.4      &       53.7          &       26.4            &         28.2       &           53.6                \\
	ReViSE                &           {\bf 93.4}               &           {\bf 93.5}      &            {\bf 92.2}       &         {\bf 65.4}         & 32.4        &           31.5        &      {\bf 68.1}                 \\ \hline
	\end{tabular}
	}
	\label{tbl:zero_recog}
	\vspace{-3mm}
	\end{table}
}

\end{document}